\documentclass{article}

\usepackage[english]{babel}
\usepackage{multirow}
\usepackage[letterpaper,top=2cm,bottom=2cm,left=3cm,right=3cm,marginparwidth=1.75cm]{geometry}

\usepackage[dvipsnames]{xcolor}
\usepackage{amsmath}
\usepackage{graphicx}
\usepackage{float}
\usepackage[colorlinks=true, allcolors=blue, breaklinks=true]{hyperref}
\usepackage[numbers,sort&compress]{natbib}
\usepackage{amsfonts}
\usepackage{authblk}  

\usepackage{url} 

\title{From Abstract to Actionable: Pairwise Shapley Values for Explainable AI}
\author[1]{Jiaxin Xu\textsuperscript{†}}
\author[2]{Hung Chau}
\author[2,*]{Angela Burden}
\affil[1]{University of Notre Dame, USA}
\affil[2]{Zillow Group, USA}
\affil[*]{Corresponding author: \texttt{angelabu@zillow.com}}

\date{} 

\begin{document}
\maketitle
\renewcommand{\thefootnote}{\textsuperscript{†}}
\footnotetext{This work was done during Jiaxin Xu’s internship at Zillow Group.}




\begin{abstract}
Explainable AI (XAI) is critical for ensuring transparency, accountability and trust in machine learning systems as black-box models are increasingly deployed within high-stakes domains. Among XAI methods, Shapley values are widely used for their fairness and consistency axioms. However, prevalent Shapley value approximation methods commonly rely on abstract baselines or computationally intensive calculations, which can limit their interpretability and scalability. To address such challenges, we propose Pairwise Shapley Values, a novel framework that grounds feature attributions in explicit, human-relatable comparisons between pairs of data instances proximal in feature space. Our method introduces pairwise reference selection combined with single-value imputation to deliver intuitive, model-agnostic explanations while significantly reducing computational overhead. Here, we demonstrate that Pairwise Shapley Values enhance interpretability across diverse regression and classification scenarios—including real estate pricing, polymer property prediction, and drug discovery datasets. We conclude that the proposed methods enable more transparent AI systems and advance the real-world applicability of XAI. 
\end{abstract}



\section{Main}

With the rapid advancement and widespread adoption of increasingly complex artificial intelligence (AI) black-box models across diverse applications, ensuring model outputs are transparent and understandable is becoming crucial, particularly when they have the potential to significantly impact people's lives. Consequently, Explainable Artificial Intelligence (XAI) has become an essential component of modern machine learning (ML)~\cite{arrieta2020explainable,das2020opportunities,adadi2018peeking}. The importance of XAI centers around three key aspects: (i) building user understanding and trust, (ii) promoting stakeholder accountability by making AI systems auditable, and (iii) supporting adherence to legal requirements and fairness standards~\cite{dwivedi2023explainable,saeed2023explainable,tjoa2020survey}. Among diverse XAI methods~\cite{arrieta2020explainable,das2020opportunities,adadi2018peeking,dwivedi2023explainable,saeed2023explainable,tjoa2020survey}, feature attribution~\cite{zhou2022feature,lundberg2018consistent,bilodeau2024impossibility} plays a pivotal role by identifying and quantifying the contribution of individual input features to a model's output, clarifying their influence on model predictions at both global and local levels. A common subset of feature attribution is additive feature attribution, where the attribution quantities sum to a specific value, such as the model’s prediction~\cite{lundberg2017unified}. Shapley values, derived from cooperative game theory~\cite{kuhn1953contributions}, are recognized for their mathematical fairness and consistency in assigning contributions to features by evaluating all possible combinations~\cite{sundararajan2020many}. To unify additive feature attribution methods, Lundberg and Lee~\cite{lundberg2017unified} introduced SHAP (SHapley Additive exPlanations), a framework that satisfies desirable properties like consistency and local accuracy, making it effective for interpreting models of increasing complexity.

\begin{figure}[ht]
\centering
\includegraphics[width=1\linewidth]{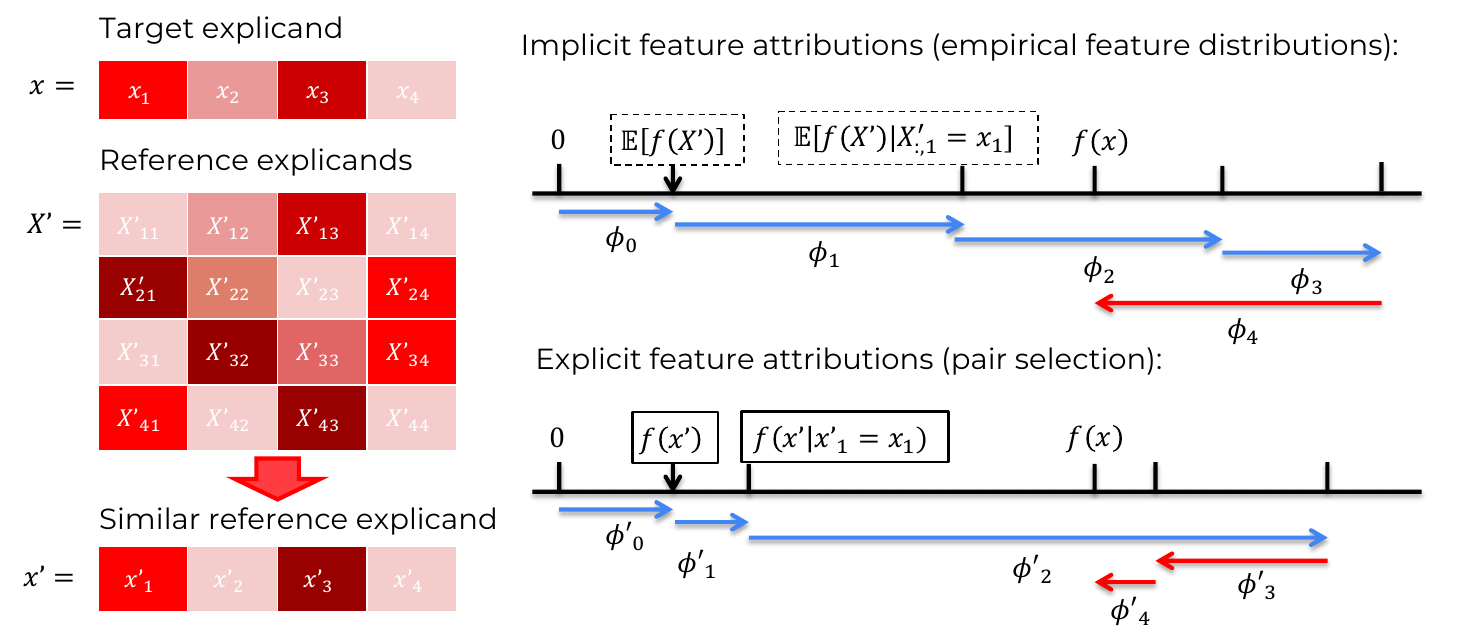}
\caption{\label{Fig1} 
Shapley value attribution comparison between traditional implicit methods using empirical feature distributions and the proposed explicit pair selection method.}


\end{figure}

Despite their theoretical appeal, Shapley value estimation methods in XAI face significant challenges in practical interpretability. Traditional explanations based on feature attribution can be complex and difficult for users to interpret. For instance, as shown in Fig. \ref{Fig1}, traditional feature attribution explanation is inherently \textit{implicit}, as it often relies on sampling from empirical feature distributions using reference explicands \( \textit{\textbf{X}}' \) (also referred to as the background dataset)~\cite{lundberg2017unified}. This opacity stems from the abstract baseline value \(\mathbb{E}[f(\textit{\textbf{X}}')]\), which represents the feature attribution when no features of target explicand \( x \) are known. Additionally, Shapley values for individual features, such as \(x_1\) (\(\phi_1\)), are computed from expectations based on the empirical feature distribution (\(\mathbb{E}[f(\textit{\textbf{X}}')\mid \textit{\textbf{X}}'_{:,1} = x_1]\)), obscuring the specific reference values. This abstraction makes it difficult to relate explanations to tangible comparisons, limiting their practical utility. Furthermore, while the Shapley value is uniquely defined by certain axioms~\cite{kuhn1953contributions}, its application in model explanation varies based on how the model, training data, and explanation context are utilized, leading to inconsistent outputs which challenges the applicability of its theoretical uniqueness~\cite{sundararajan2020many}. Additionally, Shapley value methods often face a trade-off between being true to the model (e.g., marginal Shapley values) and being true to the data (e.g., conditional Shapley values), complicating the attribution process and further reducing interpretability~\cite{chen2020true}. These challenges underscore the need for a more intuitive and accurate feature attribution method that balances model fidelity with human interpretability.

In cognitive psychology, analogical reasoning refers to the process by which individuals understand novel situations by relating them to similar and familiar ones, thereby facilitating learning and problem-solving~\cite{gentner2003analogical, goswami1996analogical}. For example, in real estate valuation, homes are valued via a comparative market analysis (CMA)~\cite{kahr2006real, yeh2018building, pagourtzi2003real}. In a CMA, a property's value (the subject home) is estimated by comparing it to recently sold properties with similar attributes (comparable homes), such as location, size, and condition. To determine a fair market price for the subject home, the sale price of the comparable home is adjusted for differences in amenities, location and time. Similarly, in drug design, the concept of a ``comparable reference'' can be applied in developing HIV treatments. Comparisons often focus on protease inhibitors sharing a common functional group essential for activity. Once this critical functional group is established, medicinal chemists can systematically modify other parts of the molecule to enhance efficacy, selectivity, and pharmacokinetics. Inspired by these human evaluative strategies, our proposed \textbf{Pairwise Shapley Value} method introduces a tangible reference point into the feature attribution process (Fig. \ref{Fig1}). By selecting a comparable baseline from the background dataset that closely matches the target explicand, the pairwise method assigns attributions to the differences in feature values between the target explicand and its reference. Pairwise Shapley values are \textit{explicit} as they ground explanations in concrete, relatable comparisons, mimicking the human decision process. Furthermore, by conditioning on certain important features in the input space, the risk of generating off-manifold data is reduced and complex feature interactions are eliminated, leading to simpler and more intuitive explanations.

In summary, our key contributions are:
\begin{itemize}
    \item Pairwise Shapley Values, a novel Shapley values-based approach that enhances human-relatable explainability through pairwise comparisons, offering greater interpretability, computational efficiency, and robustness.

    \item Extensive benchmarking of Pairwise Shapley Values against prevalent Shapley-based post-hoc explainability methods, evaluated on real-world datasets spanning real estate pricing, polymer property prediction, and drug discovery.

\end{itemize}




\section{Related Work}
\subsection{Feature Attribution Methods}

Feature attribution methods quantify individual feature contributions to an ML model’s prediction, either locally (for specific instances) or globally (aggregated across all predictions). Additive feature attribution methods, which sum feature contributions to a specific value, such as the model’s output, are notable for their clear and interpretable decomposition of predictions.
Feature attribution methods can be categorized into (i) model-agnostic methods, which generate explanations by sampling predictions from a black-box model. This includes method such as LIME (Local Interpretable Model-agnostic Explanations)~\cite{ribeiro2016should, dieber2020model, salih2024perspective,nguyen2021evaluation}, SHAP~\cite{lundberg2017unified, salih2024perspective,nguyen2021evaluation} and Shapley sampling values~\cite{strumbelj2014}; and (ii) model-dependent methods, which leverages the specific ML model architecture. This includes techniques such as using linear models, Saliency Maps~\cite{simonyan2013deep, mundhenk2019efficient, samuel2021evaluation, alqaraawi2020evaluating}, DeepLift~\cite{shrikumar2017learning, shrikumar2016not}, and Integrated Gradients~\cite{sundararajan2017axiomatic, vcik2021explaining, singh2023choose,wang2024gradient}.

\subsection{Shapley Values}

Shapley values~\cite{shapley1953value}, originating from cooperative game theory, describe how the payout of a game should be divided fairly among a coalition of players, and have several useful properties:
(i) Efficiency: individual player payouts must sum to the total payout minus the original (null player) payout,
(ii) Symmetry: when two players contribute equally to a game (over all possible player coalitions), they should be assigned the same payout,
(iii) Dummy: a player that does not influence the game should be assigned zero payout,
(iv) Additivity: For a game that includes multiple sub-games and a summed total payout, the total payout for a single player must equal the sum of their payouts for each individual sub-game.

The Shapley value for a player \( i \) is defined as:
\begin{equation}
\phi_i(v) = \sum_{S \subseteq N \setminus \{i\}} \frac{|S|! \, (|N| - |S| - 1)!}{|N|!} \left[ v(S \cup \{i\}) - v(S) \right],
\label{eq:shapleyvalues}
\end{equation}
where \( N \) is the set of all players, \( S \) is a subset of \( N \) players not containing player \( i \), \( v(S) \) represents the payout (value function) based on the subset of players \( S \), and \( |S| \) denotes the cardinality of set \( S \). Equation (\ref{eq:shapleyvalues}) calculates the weighted average of the marginal contributions of player \( i \) across all possible subsets \( S \). 

This technique has been adapted to derive explanations from black-box ML models by analogizing the model prediction as the game payout and the model features as the game players~\cite{strumbelj2014, lundberg2017unified}. However, computing Shapley values in ML settings can be complex and computationally intensive, which has spawned a variety of estimation algorithms~\cite{chen2023algorithms}.

\subsection{Feature Removal in Shapley Value Estimation}


During Shapley value estimation, feature removal simulates the absence of a feature by replacing or marginalizing its effect when computing the value function \( v(S) \)~\cite{chen2023algorithms}. The method of feature removal can significantly influence the resulting feature attribution values and their interpretation~\cite{sundararajan2020many, kumar2020problems, merrick2019games}. Two common strategies are single-value removal and distributional removal~\cite{chen2023algorithms}.

Single-value removal replaces the feature to be removed with a fixed reference value. The set of fixed reference values is often termed the baseline~\cite{sundararajan2020many}. A common choice is zero, but other values like the mean or median of the feature can also be used. \( v(S) \) is computed by setting the features not in \( S \) (denoted as \(\bar{S}\)) to their corresponding baseline values. While straightforward, this method may introduce unrealistic data instances in the Shapley value estimations that can result in misleading attributions~\cite{aas2021, Zhao_OOdistribution}. Rather than substitute a feature with a single value, distributional removal integrates over a range of possible values, considering the feature's empirical distribution. There are various ways to do this, we examine 3 common methods:



\begin{itemize}
    \item \textit{Uniform Distribution:} A feature is replaced with values sampled uniformly across its range. This technique assumes all feature values are equally probable, which might not align with the actual data distribution.
    \item \textit{Marginal Distribution:} A feature is replaced by sampling its marginal distribution. This approach breaks dependencies between features, potentially leading to implausible data points in the Shapley value estimation. The value function, \( v(S) \), is then the expected value of the model output when features not in \( S \) are marginalized independently:
    
    \begin{equation}
    v(S) = \mathbb{E}_{p(x_{\bar{S}})}[f(x_S, x_{\bar{S}})].
    \label{eq:shapleyvalues-marginal}
    \end{equation}
    
    \item \textit{Conditional Distribution:} A feature is replaced by sampling its conditional distribution relative to other features. This method preserves dependencies among features, resulting in more realistic data instances in the Shapley value estimation but at the cost of increased computational complexity. Furthermore, multicollinearity between features may cause non-zero attributions for features with no influence on the model outcome~\cite{chen2023algorithms}. The value function is the expected value given the features in \( S \), considering their dependencies:
    
    \begin{equation}
    v(S) = \mathbb{E}_{p(x_{\bar{S}} \mid x_S)}[f(x_S, x_{\bar{S}})].
    \label{eq:shapleyvalues-conditional}
    \end{equation}
\end{itemize}

Selecting the appropriate feature removal approach is important for generating meaningful and reliable Shapley value-based explanations.

\section{Pairwise Shapley Values}

Our framework integrates a ML base model with a pair selection algorithm and feature removal based on explicit pairs (the target explicand and the similar reference explicand), as illustrated in Fig. \ref{Fig1}.

\subsection{Pair Selection Algorithm} 
To provide flexibility and balance between general applicability and domain-specific precision, we employ three pair selection algorithms to identify a reference explicand for a target explicand: 
\begin{itemize}
\item  \textit{Random:} Selecting a reference explicand (\( x' \)) randomly from the background dataset (\(\textit{X}'\)). This serves as a baseline to compare against more structured selection strategies.

\item \textit{Similar:} Choosing a reference explicand that closely matches the target explicand based on common similarity metrics, such as cosine similarity, Euclidean distance, or correlation-based measures. These algorithms represent a generalizable method applicable to any use case without requiring domain knowledge.

\item \textit{Comparable:} Defining similarity based on specific application criteria. This is a specialized approach tailored to different domains, where domain knowledge is needed to identify and condition on the ``more important" factors, ensuring relevance and interpretability in context. For example, in a CMA, comparable homes are typically selected based on similarity in attributes such as sale timing, location, size, and condition~\cite{kahr2006real, yeh2018building, pagourtzi2003real}. 

\end{itemize}

\subsection{Shapley Value Estimation Based on Explicit Pairs}
\label{ExplicitPairs}

After selecting an explicit reference, we use single-value imputation to calculate the value function \( v(S) \). This involves replacing the removed features (\( \bar{S} \)) in the target instance (\( x \)) with the corresponding features in the reference instance (\( x' \)) and observing the effect on the model's prediction:

\begin{equation}
v(S) = f(x_S, x'_{\bar{S}}).
\label{eq:shapleyvalues-conditional-1}
\end{equation}

We compute the Shapley values \(\phi_i(v)\) (denoted as \(\phi'_i\) in Fig. \ref{Fig1}) using the KernelSHAP~\cite{lundberg2017unified}, a popular model-agnostic approach that approximates Shapley values by solving a weighted least squares problem. This approximation is necessary in practice because calculating exact Shapley values requires evaluating all \( 2^N \) possible feature subsets, which is computationally infeasible for models with a large number of features. However, although not implemented in this work, we note that conditioning on large subsets of features with zero variance, that we coin ``dummy pairs", where \( x_i = x'_i \), in our pair selection algorithm, allows us to omit them from the stack of evaluated feature permutations, thereby significantly decreasing the runtime of the Pairwise Shapley algorithm.

As well as adhering to the Shapley value properties (efficiency, symmetry, dummy, additivity), the pairwise method has several desirable properties not always encountered in the other Shapley value approximation methods discussed:
\begin{enumerate}
\item  \textit{Additive inverse}: Feature attributions derived by swapping the baseline and explicand have the additive inverse contribution, \(  \phi_i ( x, x' ) = -\phi_i ( x', x ) \). For example, in the home valuation problem, the contribution to the difference in home value attributed to the difference in square foot between a subject and comparable home would be \(  \phi_i \) dollars comparing subject to comparable home and \( -\phi_i \) dollars comparing comparable home to subject.
\item \textit{Dummy pairs}: Features the same in both baseline and explicand are named ``dummy pairs''. Dummy pairs are never attributed any value, but dictate the locality of the value function for evaluation. They can be considered locally independent from the remaining feature set.
\item \textit{Single feature contribution}: It follows that when differences between data pairs are isolated to a single feature, that feature (by elimination) is also independent. All changes in the model outcome will therefore be attributed to changes in that feature. 
\item  \textit{Logical attribution evaluation}: Where features are expected to contribute monotonically to the output (for example, square foot) we can easily evaluate our predictive ML model for expected behavior with the pairwise method, on real data.
\end{enumerate}

\section{Results}

\subsection{Base Machine Learning Models' Performance}
\label{BaseML}
We first train predictive ML models to ensure reliable feature attributions. Since explanations depend on model predictions, a sufficiently accurate base model is essential. Feature explanations inherently depend on the model and its predictions; hence, a sufficiently accurate predictive model is essential for meaningful and interpretable attributions. We utilized the Tree-based Pipeline Optimization Tool (TPOT)~\cite{le2020scaling}, an automated ML (AutoML) tool, to automatically select the model pipeline and hyper-parameters for our datasets. In Table \ref{tab:baseperformance} we present the performance of the base ML models on three datasets: (1) the King County home price prediction dataset~\cite{andykrause_2019_github} (referred to as ``Home''), (2) the polymer dielectric constant prediction dataset~\cite{kuenneth2021polymer} (referred to as ``Polymer''), and (3) the drug HIV replication inhibition ability prediction dataset~\cite{wu2018moleculenet} (referred to as ``Drug''). These datasets encompass both regression and classification tasks, offering a testbed for our feature attribution estimation framework. Detailed description for each dataset and task are provided in Section \ref{Methods-datasets}.

\begin{table}[ht]
\centering
\begin{tabular}{|l|c|c|c|c|}
\hline
\textbf{Dataset} & \textbf{Type} &  \textbf{Metric} & \textbf{Performance}& \textbf{Benchmark Performance} \\ \hline
Home& Regression & MdAPE & 0.106 & 0.100~\cite{moghimi2023rethinking}  \\ \hline
Polymer    & Regression & RMSE & 0.762 & 0.530~\cite{kuenneth2021polymer}\\ \hline
Drug       & Binary classification & ROC-AUC & 0.820 & 0.792~\cite{wu2018moleculenet} \\ \hline
\end{tabular}
\caption{Performance comparison of the base ML models and benchmarks across three datasets on the test data.}
\label{tab:baseperformance}
\end{table}


The base models demonstrate competitive predictive accuracy compared with commonly used benchmarks, ensuring a adequately reliable foundation for feature attribution analysis. Details of the TPOT pipeline and model training are provided in Section \ref{Methods-basemodel}, and additional visualization of results are included in the Appendix \ref{Appendix-BaseModel}.

As an illustrative example, we also summarize a protocol for iterative explainable ML model development based on feature attribution feedback and domain knowledge, using the Home dataset as an example. Details can be found in Appendix \ref{Appendix-Protocol}. This protocol highlights how explainability-driven insights can guide enhancements in model performance and interpretability, demonstrating the complementary roles of predictive accuracy and interpretability in practical ML workflows. We hope this protocol serves as a useful guideline for researchers and practitioners in domains where both high model accuracy and interpretability are essential for informed decision-making and regulatory compliance.

\subsection{A Comparison of Existing Feature Removal Methods}


Using the base ML models in Section \ref{BaseML}, we compare feature attribution results obtained from different feature removal methods in Shapley value estimation, evaluating both global (dataset-wide) and local (instance-level) explanations. We examine several established feature removal methods for Shapley value estimation, including single-value imputation methods using zeros (B0) and mean feature values (BM), uniform distribution imputation (UF), marginal distribution imputation using all training data (MA) or a K-means summary of the training data (MK), conditional distribution imputation using all training data (CA), and a model-specific method, TreeShap (TS), which is suited for tree-based ML models selected by the AutoML pipeline. Implementation details of these methods are listed in Section \ref{Implement-Existing}.

\begin{figure}[ht]
\centering
\includegraphics[width=0.7\linewidth]{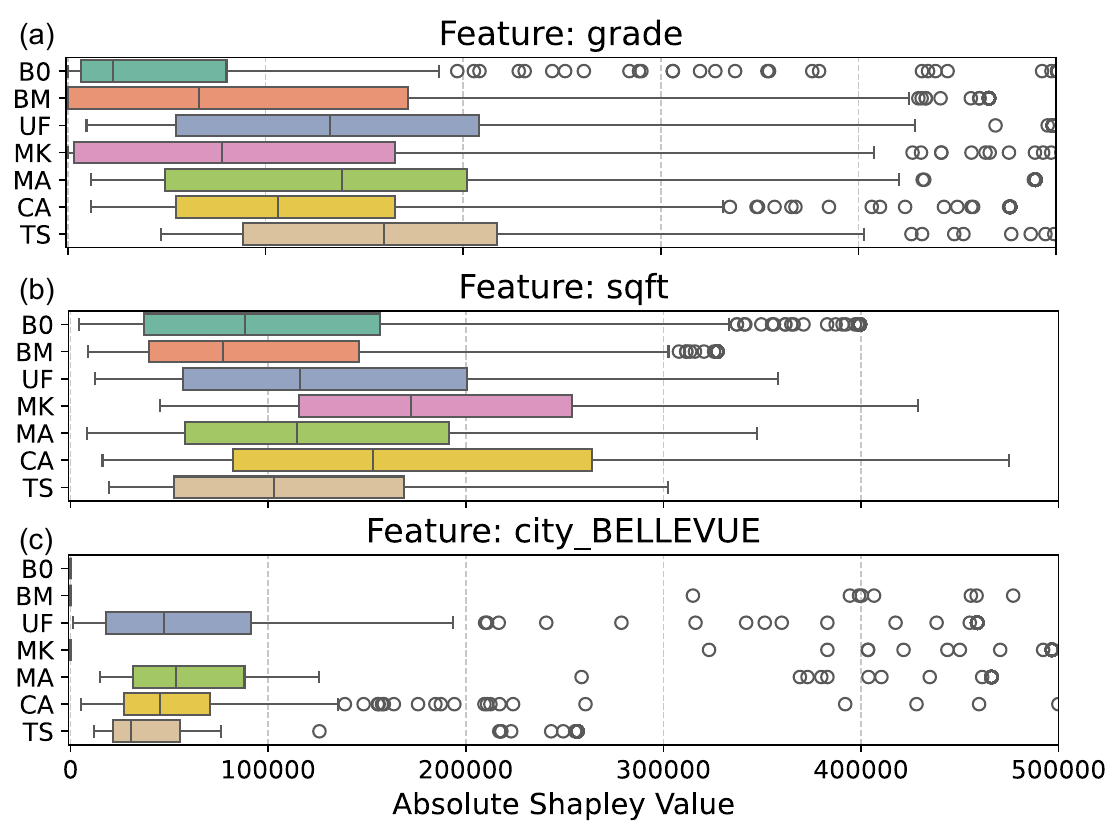}
\caption{\label{Fig4-2-feature-importance}Feature importance scores (mean absolute Shapley values) for features: (a) \texttt{grade}, (b) \texttt{city\_BELLEVUE}, and (c) \texttt{sqft}, derived from the base model on the test data of the Home dataset. The methods include single-value imputation using zeros (B0) and mean values (BM), uniform distribution imputation (UF), marginal distribution imputation using all training data (MA) or a K-means summary of the training data (MK), conditional distribution imputation using all training data (CA), and a model-specific method, TreeShap (TS).}
\end{figure}


Figure \ref{Fig4-2-feature-importance} presents the mean attributions for key features (\texttt{grade}, \texttt{sqft}, and \texttt{city\_BELLEVUE}) in the Home dataset, highlighting substantial variability across imputation methods (see initial feature importance analysis in Appendix Fig. \ref{Appendix-KingImportance}). 
For instance, among the three selected features, MA assigns significant importance to feature \texttt{grade}, while CA emphasizes feature \texttt{sqft}. Meanwhile, although MA and MK share a similar conceptual foundation, they yield markedly different results; MA assigns significant importance to the \texttt{grade} feature, whereas MK prioritizes the \texttt{sqft} feature. Similar trends appear in the Polymer and Drug datasets (Appendix Fig. \ref{Appendix-PolyDrug}), illustrating how baseline selection impacts feature importance in Shapley value estimation.

\begin{figure}[ht]
\centering
\includegraphics[width=1\linewidth]{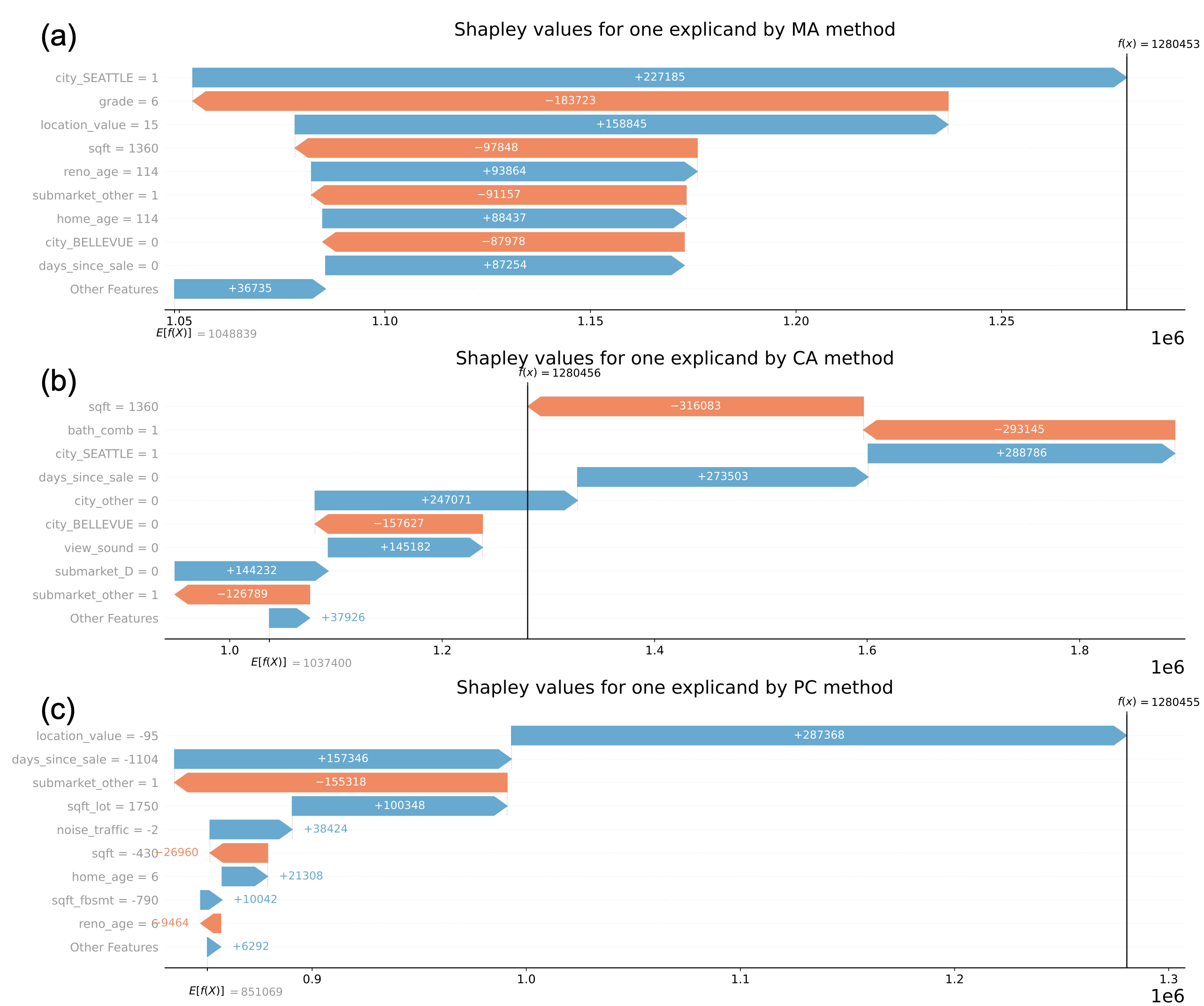}
\caption{\label{Fig4-3-local-exp}Waterfall plot of feature attributions for an explicand from the Home dataset using different methods: (a) MA, (b) CA, and (c) PC. The y-axis lists the top nine features with their values for the explicand (relative values in (c)); remaining features are aggregated as ``Other Features.'' The x-axis shows feature attributions in dollars. $f(x)$ is the model's predicted value; $E[f(X)]$ is the expected prediction based on the background data.}
\end{figure}

Figure \ref{Fig4-3-local-exp}(a)-(b) provides an example of local feature attributions for one explicand in the Home dataset. Consistent with the global results, the local attributions from MA and CA show considerable variation. In this example, MA (Fig. \ref{Fig4-3-local-exp}(a)) assigns -97k to 1360 sqft whereas CA (Fig. \ref{Fig4-3-local-exp}(b)) assigns -316k to 1360 sqft. More examples for other datasets are in Appendix \ref{Appendix-FeatureAttributionLocal} (Fig. \ref{append_local_poly} and \ref{append_local_drug}). In addition to the variation in the explanation results, a notable limitation of these imputation methods, especially distributional sampling-based methods, is their lack of direct interpretability. The reliance on imputation obscures the exact benchmark feature value against which the explicand is compared, making it challenging to derive actionable insights from the local explanations. For example, in Fig. \ref{Fig4-3-local-exp}(a), the explicand has a \texttt{grade} of 6 and a \texttt{sqft} of 1360, which are associated with approximate reductions in the predicted home price of 183k and 97k, respectively. However, we don't know the benchmark feature values of \texttt{grade} and \texttt{sqft} against which these reductions are calculated. The reference values remain obscured due to the sampling processes employed in the imputation.

To enable a fair comparison across feature removal methods and mitigate the impact of baseline variations ($E[f(X')]$ in Fig. \ref{Fig1} and $E[f(X)]$ in Fig. \ref{Fig4-3-local-exp}), we normalize feature attributions, yielding the \textit{predicted marginal change in the target variable per unit of a specific feature}. For example, in the Home dataset, this allows us to express the \texttt{sqft} feature's impact as dollars per square foot (\$/sqft), independent of baseline selection. Normalization is performed by computing differences in Shapley values and feature values across random explicand pairs:


\begin{equation}
f(x^i) = \phi_0 +  \sum_{k=1}^{n}\phi_k^i
\label{eq:shapley-decomposition}
\end{equation}

\begin{equation}
\text{norm}_{k}^{ij} = \frac{\phi_k^i - \phi_k^j}{x_k^i - x_k^j},
\label{eq:normalized-difference}
\end{equation}
where $x^i$ is an explicand with $n$ features and the $k^{th}$ feature value is $x^i_k$; $f(x^i)$ is the ML predicted value of explicand $x^i$; $\phi_0$ is the baseline value of the feature removal method, $\phi_k^i$ is the Shapley value of the $k^{th}$ feature in explicand $x^i$, and $\text{norm}^{ij}_k$ is the normalized Shapley value of feature $k$ between two explicands $x^i$ and $x^j$.





\begin{figure}[ht]
\centering
\includegraphics[width=0.8\linewidth]{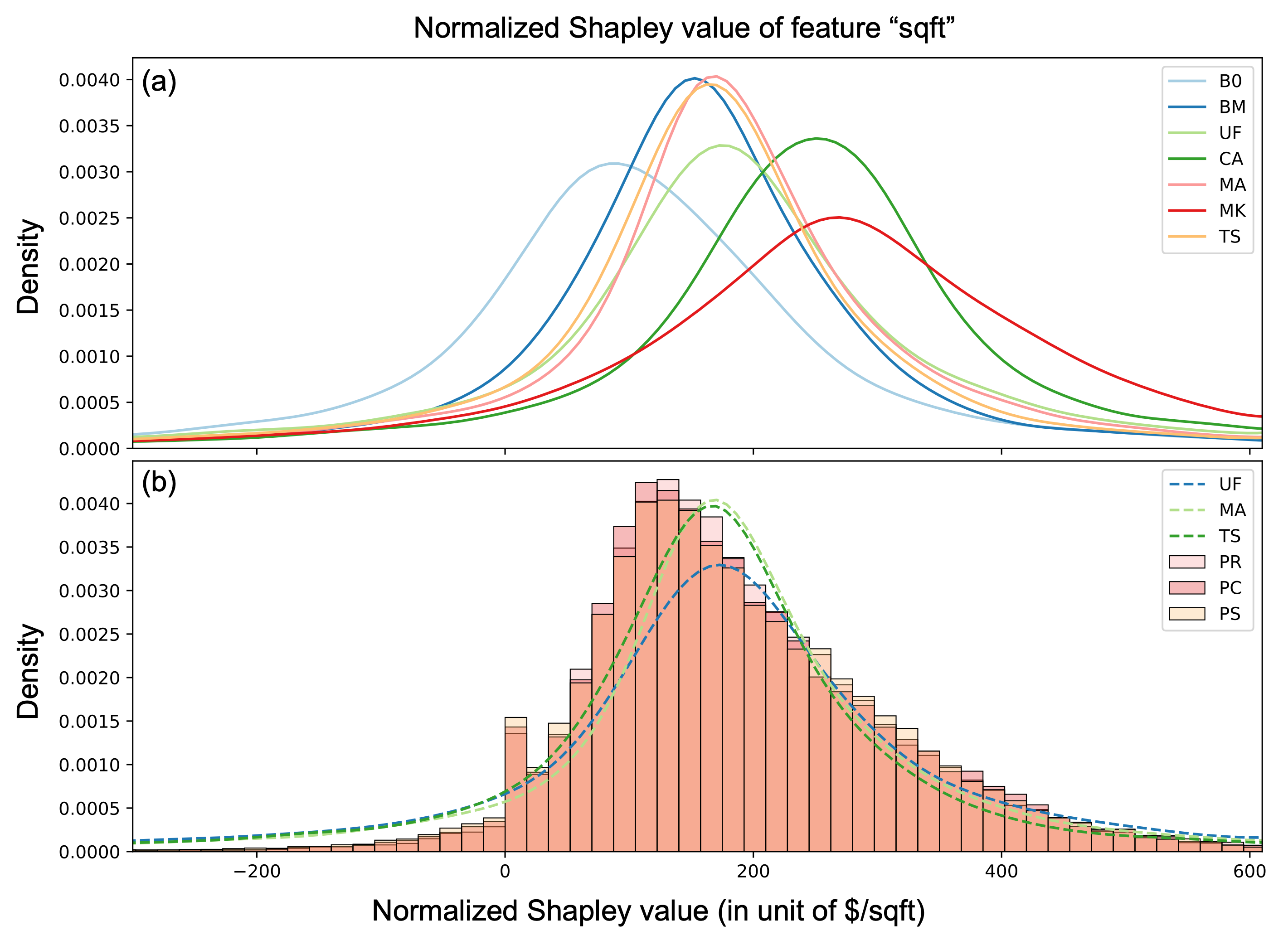}
\caption{\label{Fig4-4-normalized}Distribution of normalized Shapley values for feature \texttt{sqft} in the Home dataset for different methods. (a) Distribution plot for seven existing non-pairwise methods; (b) Distribution plot for three selected existing non-pairwise methods (UF, MA, TS) and three pairwise methods with different similarity algorithms (PR, PC, PS).}
\end{figure}

Using feature \texttt{sqft} in the Home dataset as an example, the resulting distribution of $\text{norm}_{\texttt{sqft}}$ from 500 pairs are shown in Fig. \ref{Fig4-4-normalized}(a). The mean and standard deviation across methods are summarized in Table \ref{tab:sqft_summary_statistics}. A Kolmogorov–Smirnov (KS) test (using MA as the reference) confirms that normalized Shapley distributions significantly differ (p-values $<$ 0.01, see Table \ref{tab:sqft_summary_statistics}), underscoring the variability across feature removal techniques.



\begin{table}[ht]
\centering
\renewcommand{\arraystretch}{1.2}
\begin{tabular}{|l|c|c|c|r|r|r|r|}
\hline
\textbf{Method} & \textbf{Mean} & \textbf{Std} & \textbf{Skew} & \multicolumn{4}{c|}{\textbf{p-value}} \\ \cline{5-8}
 & & & & \textbf{MA} & \textbf{PC} & \textbf{PS} & \textbf{PR} \\ \hline
\textbf{B0} & 94.74 & 249.76 & -0.56 & 0.0 & - & - & - \\
\textbf{BM} & 149.43 & 213.63 & -0.99 & 1.1e-165 & - & - & - \\
\textbf{UF} & 169.82 & 264.3 & -0.93 & 3.1e-26 & 1.4e-146 & 3.5e-151 & 1.4e-180 \\
\textbf{CA} & 233.55 & 244.03 & -1.35 & 0.0 & - & - & - \\
\textbf{MA} & 172.08 & 238.74 & -1.08 & - & 1.9e-102 & 2.5e-92 & 6.4e-115 \\
\textbf{MK} & 259.86 & 269.97 & -1.24 & 0.0 & - & - & - \\
\textbf{TS} & 157.95 & 235.35 & -1.03 & 7.8e-41 & 8.6e-111 & 8.8e-100 & 6.1e-124 \\
\hline
\textbf{PC} & 188.96 & 138.72 & 0.33 & - & - & 0.0184 & 0.00041 \\
\textbf{PS} & 188.91 & 138.87 & 0.38 & - & 0.0184 & - & 0.00016 \\
\textbf{PR} & 190.5 & 135.21 & 0.47 & - & 0.00041 & 0.00016 & - \\
\hline
\end{tabular}
\caption{Summary statistics and Kolmogorov–Smirnov (KS) test results for the distribution of normalized \texttt{sqft} feature attribution from different methods. The p-values indicate statistical comparisons, where MA, PC, PS, and PR serve as the reference distributions.}
\label{tab:sqft_summary_statistics}
\end{table}

\subsection{Pairwise Shapley: Enhanced Interpretability}

We compute the feature attribution values with the pairwise framework using two different pair selection strategies. First, we reuse the same random pairs of explicands from the prior analysis in normalized Shapley values, denoted as ``PR'' (Pairwise-Random). Second, we devise two similarity-based algorithms to select pairs of comparable explicands for pairwise computation, focusing on identifying explicands with close values for key attributes, denoted as ``PS'' (Pairwise-Similar) and ``PC'' (Pairwise-Comparable).

Pairwise Shapley values show more intuitive explanations by controlling the background data and using informative background feature values, particularly in local explanations. For example, using PC method on the Home dataset, in Fig. \ref{Fig4-3-local-exp}(c), the explanation shows that a target home with 430 square feet less than the reference (a comparable home) results in a \$27k decrease in the final home price prediction. Similarly, the target home having a traffic noise level 2 units lower than the comparable reference leads to a \$38k increase in the predicted home price. This explanation closely mirrors the appraisal process in real-world home valuation, making the explanations more intuitive compared to the non-pairwise methods shown in Fig. \ref{Fig4-3-local-exp} (a)-(b). Similar benefits are observed in Polymer and Drug datasets (Appendix \ref{Appendix-FeatureAttributionLocal}).

\subsection{Pairwise Shapley: Improved Model Insights}
Beyond more intuitive explanations, Pairwise Shapley allows us to derive additional insights into model behavior. In this section we show how we can derive model insights by (i) comparing the distributions of ``normalized'' attribution values (ii) measuring the rate of sign matching between attribution values and feature differences, and (iii) observing global attribution values as a function of feature (difference) value.

\subsubsection{Normalized Shapley value}
\label{Norm_sqft}

Different from Eq.~\ref{eq:normalized-difference}, we compute the normalized Shapley values of pairwise methods directly from explicand pairs as:


\begin{equation}
\text{norm}^{ij}_k|_{\text{pairwise}} = \frac{\phi^{ij}_k}{x^i_k - x^j_k},
\label{eq:normalized-difference-pair}
\end{equation}
where $x^i$ is the target explicand and $x^j$ is the reference explicand in pairwise method; $\phi^{ij}_k$ is the Pairwise Shapley value of feature $k$ in explicand $x^i$ compared to explicand $x^j$.

Figure \ref{Fig4-4-normalized}(b) and Table \ref{tab:sqft_summary_statistics} show the distribution and statistics of normalized Shapley values of feature \texttt{sqft} in the Home dataset for pairwise and non-pairwise methods. Pairwise methods recover a positively skewed distribution of normalized \texttt{sqft} attribution, aligning with our domain knowledge, while non-pairwise methods produce negative skew. A KS test finds no significant differences among pairwise distributions, reinforcing their robustness. Additionally, the pairwise approach requires no extra post-processing, as normalization is inherently built into the attribution process.



\subsubsection{Directionality of feature attributions}
\label{Monotonicity}

\begin{figure}[ht]
\centering
\includegraphics[width=1.0\linewidth]{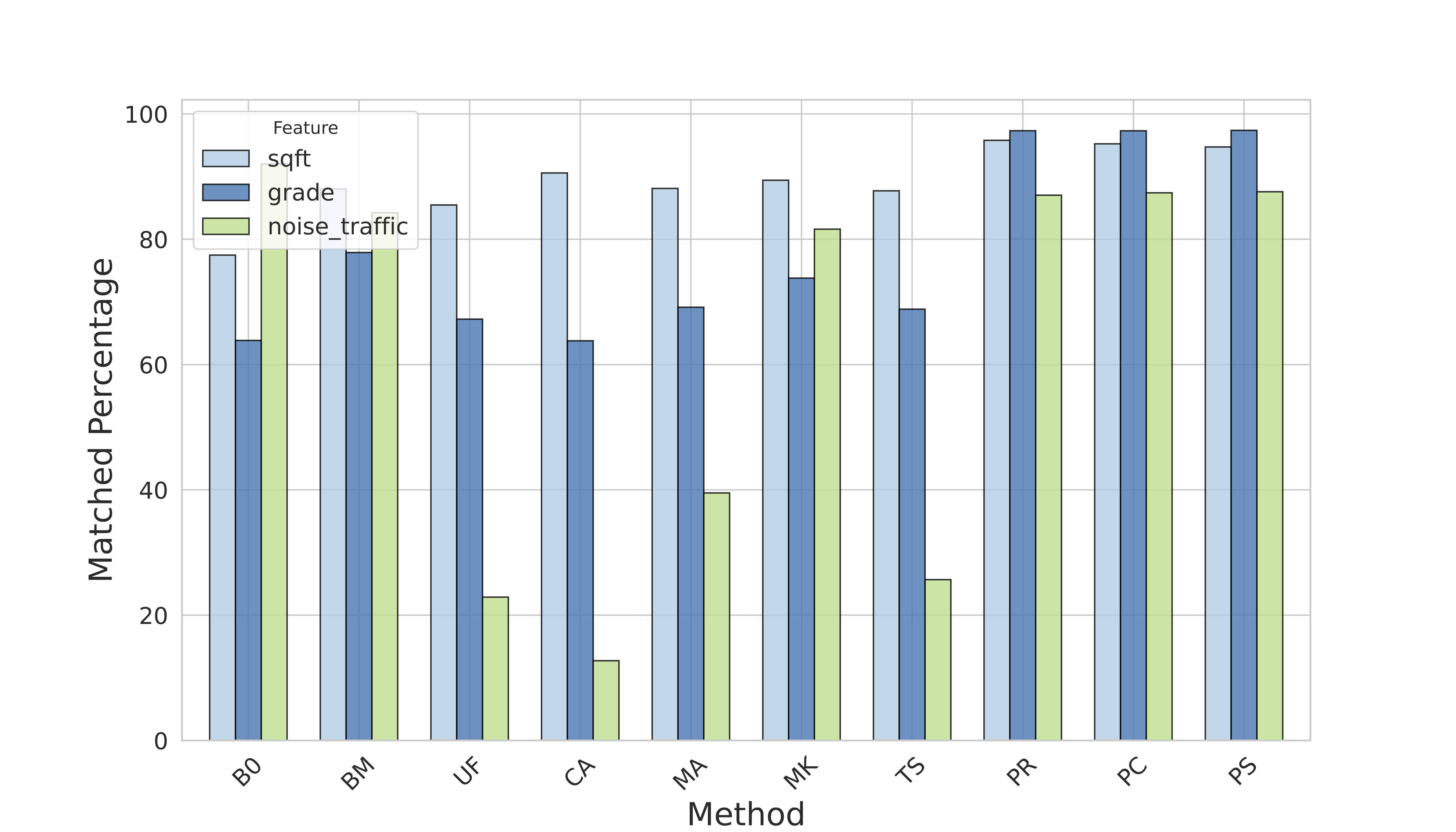}
\caption{\label{Fig4-6-mono-matchsign}Monotonicity measure (matched percentage) of Shapley value estimates across different methods for three features: \texttt{sqft}, \texttt{grade}, and \texttt{noise\_traffic}.
}

\end{figure}

For key features, such as \texttt{sqft}, \texttt{grade}, and \texttt{noise\_traffic}, we evaluate the attribution values from a monotonicity perspective to show how the pairwise method allows us to evaluate logical attributions. In home valuation, while the exact dollar impact of certain features (e.g., bedrooms, view level, traffic noise) are unknown and/or subjective, the direction of their impacts can be hypothesized. For example, \texttt{sqft} and \texttt{grade} should positively impact price (i.e., higher feature values lead to higher prices), whereas \texttt{noise\_traffic} is expected to have a negative impact (see Appendix Fig. \ref{Appendix-KingImportance}). To assess the expected directionality, we compute the percentage of $\text{norm}_{k}^{ij}$ values that satisfy the following criteria: (1) $\text{norm}_{k}^{ij} > 0$ for features positively correlated with price (e.g., \texttt{sqft} and \texttt{grade}), (2) $\text{norm}_{k}^{ij} < 0$ for negatively correlated features (e.g., \texttt{noise\_traffic}), and (3) $\text{norm}_{k}^{ij} = 0$ for $x^i_k = x^j_k$ (dummy pairs). As shown in Fig. \ref{Fig4-6-mono-matchsign}, pairwise methods have the highest monotonicity rates-defined as the matched percentage-across all three features. By evaluating the monotonicity rate of these features, we can deduce if our model is behaving in an expected manner. The high monotonicy rate of these three features in the pairwise method suggests our model explanations align with human expectations and that our model is learning intuitive correlations.

\subsubsection{Visualizing global attributions}
\label{global_Visualizing}
\begin{figure}[ht]
\centering
\includegraphics[width=1\linewidth]{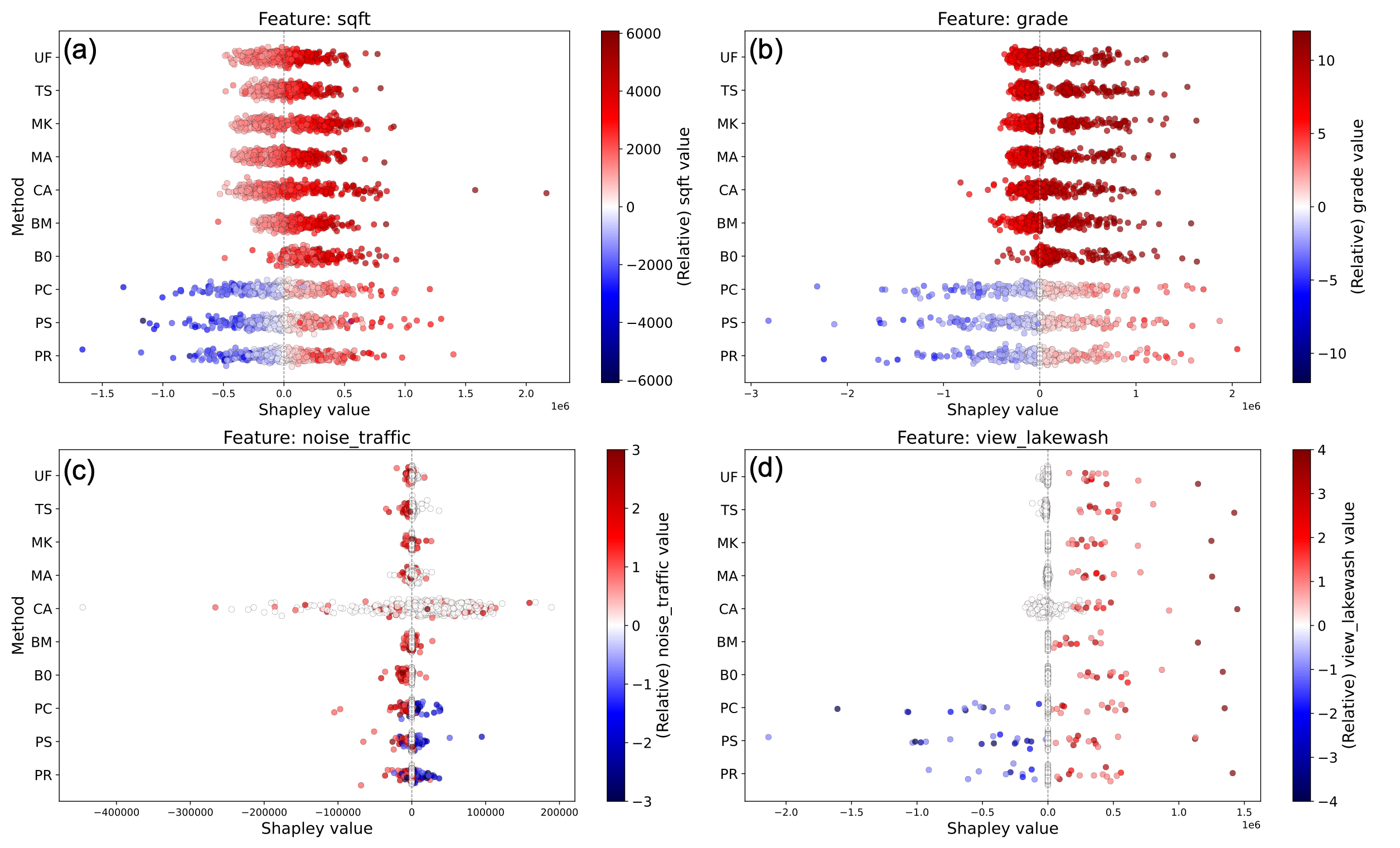}
\caption{\label{Fig4-5-beeswarm}Beeswarm plots showing the relationship between Shapley values and (relative) feature values for several methods on important features in Home dataset: (a) \texttt{sqft}, (b) \texttt{grade}, (c) \texttt{noise\_traffic}, and (d) \texttt{view\_lakewash}. The x-axis represents Shapley values (in dollars), and the color bar indicates the feature value for non-pairwise methods and the relative feature value for pairwise methods.}
\end{figure}

Figure \ref{Fig4-5-beeswarm} illustrates how the pairwise method provides a more informative representation of global model behavior. In Fig. \ref{Fig4-5-beeswarm}(a), Pairwise Shapley attributions for \texttt{sqft} increase when the \texttt{sqft} of the target home exceeds that of the reference, aligning with the real-world intuition. In contrast, non-pairwise methods assign Shapley values based on absolute feature values, where \texttt{sqft} is always positive but may represent either an increase or a decrease in predicted home value. This discrepancy becomes particularly noticeable when the feature's monotonic correlation with the target variable is weak, such as \texttt{grade}, \texttt{noise\_traffic}, and \texttt{view\_lakewash}, as depicted in Figs. \ref{Fig4-5-beeswarm}(b)-(d). In pairwise methods, the relative feature values allow us to clearly infer how an increase or decrease in a feature (e.g., a reduction in traffic noise or the presence of a certain view) impacts the predicted home price. Conversely, non-pairwise methods fail to provide such clear interpretations, as they rely on absolute feature values that do not directly reflect the relative change or its contextual meaning, making the results less actionable with reduced generalization.

\subsection{Pairwise Shapley: Feature Independence}

\subsubsection{Dummy Pair}



\begin{figure}[ht]
\centering
\includegraphics[width=0.9\linewidth]{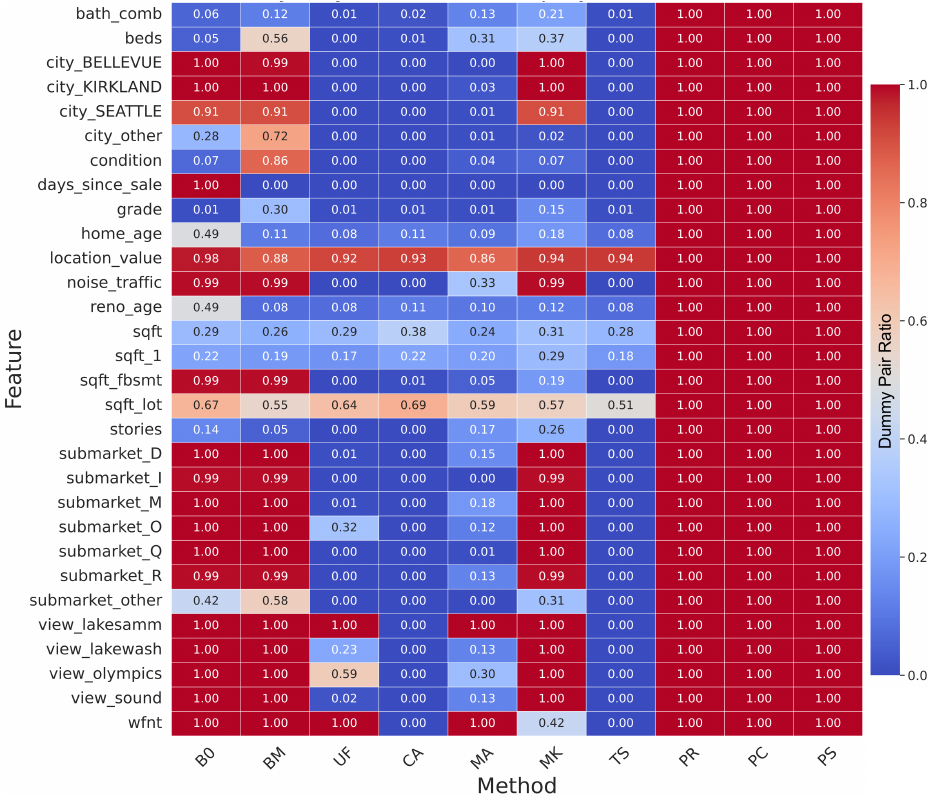}
\caption{\label{Fig4-7-dummy-player}Heatmap of dummy pair ratios across features and methods for the Home dataset.}
\end{figure}

Pairwise methods always assign zero attribution to dummy features, as the relative feature value is zero. The dummy pair property is valuable as it enables conditioning on specific features to exclude them from explanations (and computations) by selecting pairs of instances with multiple identical feature values. For example, in home valuation, we can compare similar homes within the same locality to isolate local market effects. In the context of polymers or drug molecules, we can focus on the effects of specific functional groups or motifs while excluding shared structural features that are not of interest. In that case, we can condition on shared backbone structures to better isolate and analyze the influence of substituents or side groups on target properties. Figure. \ref{Fig4-7-dummy-player} shows the dummy pair ratio (the ratio of $\text{norm}^{ij}_k = 0$ when $x^i_k = x^j_k$) of each method. Non-pairwise methods fail to consistently assign zero attribution to dummy pairs, with CA exhibiting particularly low ratios. This is expected, as conditional Shapley methods assume dependencies among features based on the original data distribution. In contrast, the pairwise method always assigns zero value to a dummy feature pair.

\subsubsection{Single feature independence}

The dummy pair property ensures that when only one feature differs between two explicands, the entire prediction difference is attributed to that feature.
For instance, when explaining the predicted value of a target home using a comparable home, if the two homes are otherwise identical, the full difference in value should be attributed to the one different feature. However, in non-pairwise methods this does not hold true.

To evaluate single feature independence in the context of multicollinearity, we conduct a feature perturbation test by slightly modifying a single feature and analyzing how attributions change. Intuitively, if all other features are held constant, and the data points are close in feature space, the full difference in the prediction should be attributed to the perturbed feature.

For this evaluation, we focus on \texttt{sqft} in the Home data due to its continuous nature, minimizing the risk of generating out-of-manifold data for the trained ML model. Synthetic data pairs are created by perturbing the \texttt{sqft} of 500 test instances in steps of $\pm$50 \texttt{sqft} (range: $-$250 to 250). We compare the prediction change with the corresponding Shapley value change to determine whether the attribution is correctly assigned to \texttt{sqft} alone. In pairwise method, the target home is the original test instance, and the reference home is the same instance with a perturbed \texttt{sqft}. In non-pairwise methods, we calculate the feature attribution for both the original and perturbed data points and then compare the change between the original and perturbed data.

\begin{figure}[ht]
\centering
\includegraphics[width=1.0\linewidth]{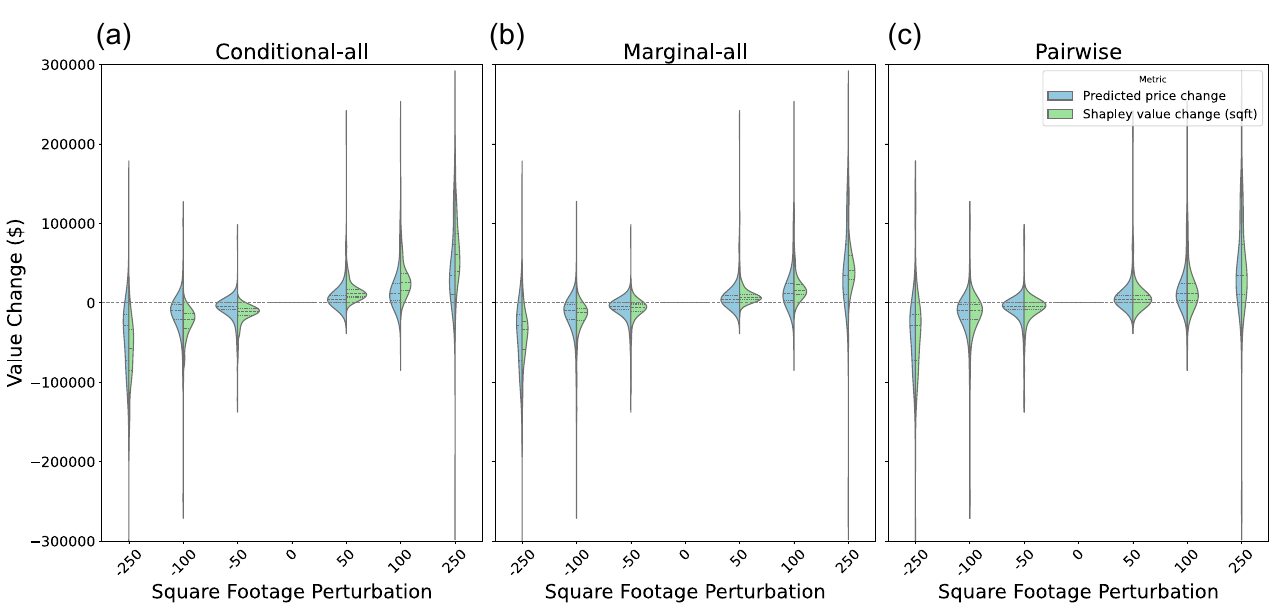}
\caption{\label{Fig4-8-perturb}Distribution of predicted price change (blue, LHS violin) and Shapley value change for \texttt{sqft} feature (green, RHS violin) in the Home dataset under feature perturbation test. (a) Conditional-all method, (b) Marginal-all method, and (c) Pairwise method. In the Pairwise method, the predicted price change and Shapley value change are always the same, leading to the identical distributions.}
\end{figure}

Figure \ref{Fig4-8-perturb} shows a split violin plot, with the left side depicting prediction differences distribution and the right side showing Shapley value differences distribution for \texttt{sqft}. In non-pairwise methods (e.g., conditional-all and marginal-all), the left and right distributions are different, particularly for the conditional-all method, where prediction changes are spread across multiple features. In contrast, the pairwise method maintains identical distributions, confirming the full prediction difference is attributed to the perturbed feature. This highlights the superiority of the pairwise method, which reliably attributes the entire change to the perturbed feature, ensuring interpretability.


\subsection{The Role of Similarity in Pairwise Shapley}

\begin{figure}[ht]
\centering
\includegraphics[width=0.5\linewidth]{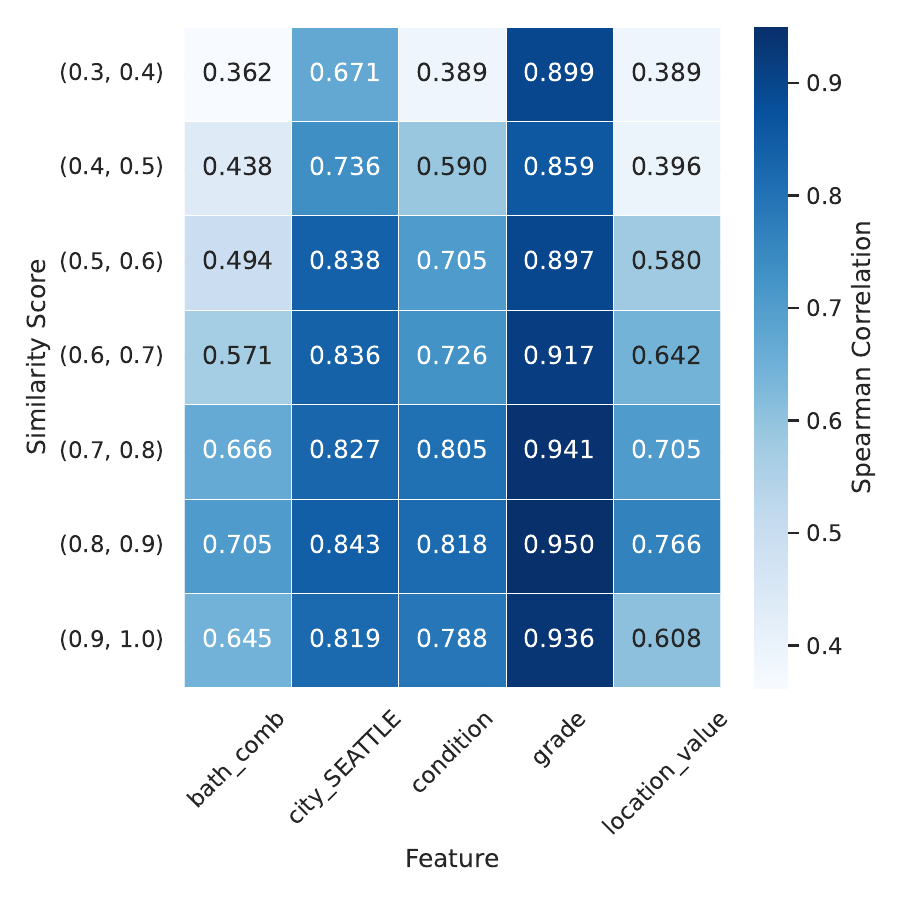}
\caption{\label{Fig4-9-simscore}Heatmap of monotonicity scores (measured using Spearman correlation) as a function of similarity scores for selected positive features in the Home dataset.}
\end{figure}

Using similar data instances as reference explicands in the pairwise method offers several key advantages. First, it reduces the likelihood of off-manifold inputs, which can distort feature attributions due to unrealistic data instances unsupported by training data~\cite{Zhao_OOdistribution}. Second, it simplifies explanations by conditioning on matched features, effectively removing them from the attribution process. To evaluate the impact of similarity on Pairwise Shapley values, we examine how monotonicity rates (as defined in Section \ref{Monotonicity}) change with explicand-reference similarity in the Home dataset. Figure \ref{Fig4-9-simscore} shows a heatmap of monotonicity scores (measured by Spearman correlation) as a function of similarity scores for several positive features in the Home dataset. The result shows that as the similarity between the target and reference explicand increases, the monotonicity rate of feature attributions generally increases. This suggests that higher similarity between data points can lead to more intuitive explanations using the pairwise method, assuming the base model has learned the expected relationship between that feature and the outcome.

\subsection{Runtime Comparison}
We compare the computational time of various explanation methods for generating Shapley values, for one single explicand from the Home dataset. As shown in Table \ref{tab:computational-time}, the Pairwise method is the most efficient, requiring only 0.08 seconds, even without implementing the dummy pair speed-up described in Section \ref{ExplicitPairs}, which could further reduce runtime by many orders of magnitude. This efficiency stems from its reliance on simple single-value imputation and the ability to perform similarity calculations independently of the Shapley value computation. In contrast, marginal and conditional methods are significantly slower due to complex imputations. Detailed settings for each method are in Section \ref{Implement-Existing}. The Pairwise method’s efficiency, combined with its intuitive explanations, makes it better-suited for larger datasets and real-time applications.

\begin{table}[ht]
\centering
\caption{Comparison of computational time across methods for explaining a single explicand. Reported values are the mean over 50 independent runs, with standard deviations in parentheses.}
\label{tab:computational-time}
\begin{tabular}{|l|c|}
\hline
\textbf{Method}       & \textbf{Computational Time (s)} \\ \hline
B0             & 0.08 (0.01)                            \\ \hline
UF              & 0.21 (0.00)                            \\ \hline
CA          & 1.71 (0.21)                          \\ \hline
MA         & 1.83 (0.12)                           \\ \hline
MK         & 2.74 (0.05)                            \\ \hline
TS             & 0.19 (0.00)                            \\ \hline
Pairwise             & 0.08 (0.01)                            \\ \hline
\end{tabular}
\end{table}

\section{Discussion}

In this work, we introduced Pairwise Shapley Values, a novel method for generating intuitive and human-interpretable model explanations by leveraging explicit comparisons between proximal data points. Unlike traditional Shapley value estimation techniques that rely on abstract distributional baselines, our method directly attributes changes in model predictions to observed feature differences, enhancing interpretability across diverse domains. By ensuring that identical feature values in explicand-baseline pairs contribute no attribution, Pairwise Shapley Values maintain local independence, enabling more precise feature conditioning and computational efficiency. This property simplifies explanations by eliminating unnecessary perturbations and ensuring that isolated feature differences fully account for prediction variations. Furthermore, our approach is robust to the pair selection routine, demonstrating consistent attributions while significantly reducing computational overhead compared to traditional methods. These characteristics make Pairwise Shapley Values well-suited for tasks requiring transparent, contextually meaningful explanations, such as real estate valuation, materials discovery, and credit risk assessment.

A key distinction of our work is its contrast with the Formulate, Approximate, Explain (FAE) framework proposed by Merrick and Taly~\cite{merrick2019games}. Both approaches are motivated by the same theoretical concept of using reference points for feature attribution, but FAE constructs attributions using a distribution of reference points with a particular property, whereas our method employs a single-point pairwise comparison. This design choice enables Pairwise Shapley Values to provide more interpretable, instance-specific explanations, avoiding reliance on distributional assumptions. One advantage of FAE is its ability to derive uncertainty estimates for attributions, which is outside the scope of our work. Theoretically, our attributions are point values, and any uncertainty in attributions would propagate from the upstream model prediction uncertainty.

Pairwise Shapley Values introduce computational efficiency benefits, particularly through dummy pair independence, where features identical in both explicand and baseline receive zero attribution. This property allows us to: (i) condition on specific features, thereby reducing explanation complexity, and (ii) achieve significant runtime improvements by eliminating redundant computations. Specifically, removing these features ($n_c$) from the perturbation routine reduces computational complexity from $2^n$ to $2^{n-n_c}$, a critical bottleneck in many existing Shapley value algorithms~\cite{olsen2024comparative}. This structured feature selection not only enhances scalability but also ensures that attributions remain relevant to the given prediction task.

A crucial consideration in Shapley value estimation is feature dependence, especially when handling missing data. Traditional approaches model missing features using either marginal or conditional distributions, each of which introduces challenges. Marginal distribution-based methods assume feature independence, potentially creating unrealistic data points, while conditional distribution-based methods retain correlations among missing and observed features, leading to unexpected attributions for non-contributing features~\cite{aas2021, merrick2019games, frye2020}. Our dummy pair-based approach ensures that missing features are locally independent and do not contribute directly to attributions. However, missing values still influence the locality of the value function for evaluation. Our pair selection algorithm mitigates the risk of generating non-representative data by identifying data points close in feature space, though this effectiveness depends on dataset characteristics. In high-dimensional or sparse datasets, finding suitable reference points may be challenging, and outliers may lack appropriate pairwise comparisons, potentially reducing explanation reliability.

Another key limitation of Pairwise Shapley Values is that, like other feature attribution methods, it captures correlations rather than causal relationships. Marginal and conditional Shapley values fail to explicitly account for causality, distributing attributions among correlated features in ways that may not reflect their actual causal impact on the target variable~\cite{chen2023algorithms}. Similarly, our pairwise method removes features based on their relationship to a neighboring reference but does not model causal mechanisms. Causal inference-based Shapley methods, such as Causal Shapley Values~\cite{Heskes2020}, Asymmetric Shapley Values~\cite{frye2020b}, and Shapley Flow~\cite{pmlr-v130-wang21b}, incorporate directed causal graphs to produce more meaningful attributions. However, these methods require prior knowledge of the causal structure, which is often unavailable in real-world applications, limiting their practicality. Instead, our method provides an interpretable way to assess how model predictions change based on observed feature variations, serving as a valuable tool for model validation and debugging.

Finally, our experiments have primarily focused on tabular data, but the Pairwise Shapley framework can be extended to structured data types, including images, text, and graphs, by defining similarity measures for pair selection. For instance, in computer vision, a reference image with similar features could be selected for comparison, while in natural language processing, semantically similar text instances could serve as baselines. Future work will explore the adaptation of Pairwise Shapley Values to these domains, further expanding its utility in  XAI.

\section{Methods} 

\subsection{Datasets and Preprocessing}
\label{Methods-datasets}
This section provides detailed descriptions of the three datasets used for evaluating the effectiveness of the Pairwise Shapley Values method across both regression and classification tasks.

\subsubsection{Home Dataset}
The Home dataset contains over 560,000 records of single-family and townhome sales in King County, Washington, from 1999 to December 2023~\cite{andykrause_2019_github}. The task is to predict the sale price of a home, making it a regression problem. This dataset includes various features categorized into general information (e.g., transactional and temporal details), geographic information (e.g., locational and zoning data), physical attributes (e.g., property size, layout, and quality), and surrounding features (e.g., environmental factors and neighborhood amenities). These diverse features provide a comprehensive testbed for evaluating feature attribution methods. This dataset is divided into training, validation, and testing sets based on the sale date of the properties: the training set includes sales from January 2020 to October 2023, the validation set covers sales in November 2023, and the testing set consists of sales in December 2023.

\subsubsection{Polymer Dataset}
The Polymer dataset focuses on predicting the dielectric constant of polymers, a critical property for materials used in electronic and energy storage applications~\cite{kuenneth2021polymer}. The dielectric constant reflects a material’s ability to store electrical energy, and accurate prediction of this property is crucial for the design of advanced polymeric materials. The labels in this dataset are obtained from Density Functional Theory (DFT) calculations, which are widely used quantum mechanical modeling methods for predicting the electronic structure of molecules and materials. Each polymer is represented using its Simplified Molecular Input Line Entry System (SMILES) string, a standardized text-based format for describing chemical structures. To convert the SMILES representation into numerical vectors suitable for machine learning, we use MACCS fingerprints, which are binary vectors where each bit represents the presence or absence of a specific substructure in the molecule. Importantly, each bit in the MACCS fingerprint is interpretable, enabling direct insights into which molecular substructures influence the dielectric constant. This dataset is randomly divided into training and testing sets with a 4:1 ratio.

\subsubsection{Drug Dataset}
The Drug dataset, sourced from MoleculeNet~\cite{wu2018moleculenet}, is a well-known benchmark for molecular ML. It originates from the Drug Therapeutics Program (DTP) AIDS Antiviral Screen, which tested the ability of over 40,000 compounds to inhibit HIV replication. The screening results were categorized into three labels: \textit{confirmed inactive} (CI), \textit{confirmed active} (CA), and \textit{confirmed moderately active} (CM). For this study, we used it as a binary classification task by combining the \textit{active} (CA) and \textit{moderately active} (CM) labels into a single \textit{active} class, resulting in a task to classify compounds as either \textit{inactive} or \textit{active}. Similar to the Polymer dataset, each compound is represented using its SMILES string. MACCS fingerprints are used to transform the SMILES representation into numerical vectors, with each bit corresponding to the presence or absence of specific chemical substructures. This binary encoding ensures interpretability and allows for systematic analysis of which substructures contribute to HIV replication inhibition. This dataset is stratified into training and testing sets using a 4:1 ratio to preserve the proportion of active and inactive compounds in both sets.

\subsection{Base Model Training}
\label{Methods-basemodel}

To evaluate the effectiveness of Pairwise Shapley as a model-agnostic feature attribution method, we trained base ML models on three datasets: Home, Polymer, and Drug. TPOT, an automated machine learning framework that utilizes genetic programming to optimize model pipelines~\cite{le2020scaling}, was employed for model selection and hyperparameter tuning. For each dataset, TPOT was configured with a population size of 30 and run for 3 generations. The final selected model for the Home dataset was a Random Forest regression model with 100 trees, a minimum of 15 samples per leaf node, and a minimum of 12 samples required for a split. For the Polymer dataset, the best-performing model was an XGBoost regression model with a learning rate of 0.1, a maximum tree depth of 8, a minimum child weight of 4, and a subsample ratio of 0.4. For the Drug dataset, the selected model was a Random Forest classification model consisting of 100 trees, requiring a minimum of 3 samples per leaf node, a minimum of 5 samples for splitting, and considering 20\% of the features when determining splits. While the primary objective was not to develop the most accurate models possible, the selected models provided sufficient predictive capability to facilitate an effective evaluation of the proposed Pairwise Shapley feature attribution method.

\subsection{Implementation of Existing Feature Removal Methods}
\label{Implement-Existing}

To estimate Shapley values for ML model feature attribution, we utilize multiple established feature removal methods, each corresponding to a distinct imputation strategy for missing features. We use the SHAP package~\cite{lundberg2017unified} with KernelExplainer for baseline imputation, where missing feature values are either replaced with the median values of the training dataset (BM) or set to zero (B0). For uniform distribution imputation (UF), we apply SamplingExplainer from SHAP, selecting 100 samples from the training dataset with a fixed random seed. Marginal distribution imputation (MA, MK) is implemented using KernelExplainer, with MA sampling 100 instances from the full training dataset and MK applying a k-means clustering summary (k=10) to create representative feature samples. Conditional distribution imputation (CA) is handled using the \texttt{shapr} library~\cite{sellereite2020shapr}, specifically applying the empirical approach, with 100 training samples, 100 feature combinations, and batch processing in 10 batches. For tree-based models, we use TreeExplainer from SHAP with the tree-path-dependent feature perturbation strategy. All experimentsmin this work were conducted on a high-performance computing node equipped with a 32-core Intel Xeon r6i processor and 256 GiB of RAM.

\section*{Acknowledgments}  
We would like to thank Dr. Simon Nilsson for valuable insights, discussions, comments, and suggestions.

\section*{Code and Data Availability}

The code and datasets used in this study are available~\href{https://github.com/Jiaxin-Xu/PairwiseShapley.git}{here}\footnote{\url{https://github.com/Jiaxin-Xu/PairwiseShapley.git}}.

\bibliographystyle{unsrt}

\bibliography{main}

\appendix

\section{Additional Results}
\subsection{Base Machine Learning Models Performance}
\label{Appendix-BaseModel}
\begin{figure}[H]
\centering
\includegraphics[width=1\linewidth]{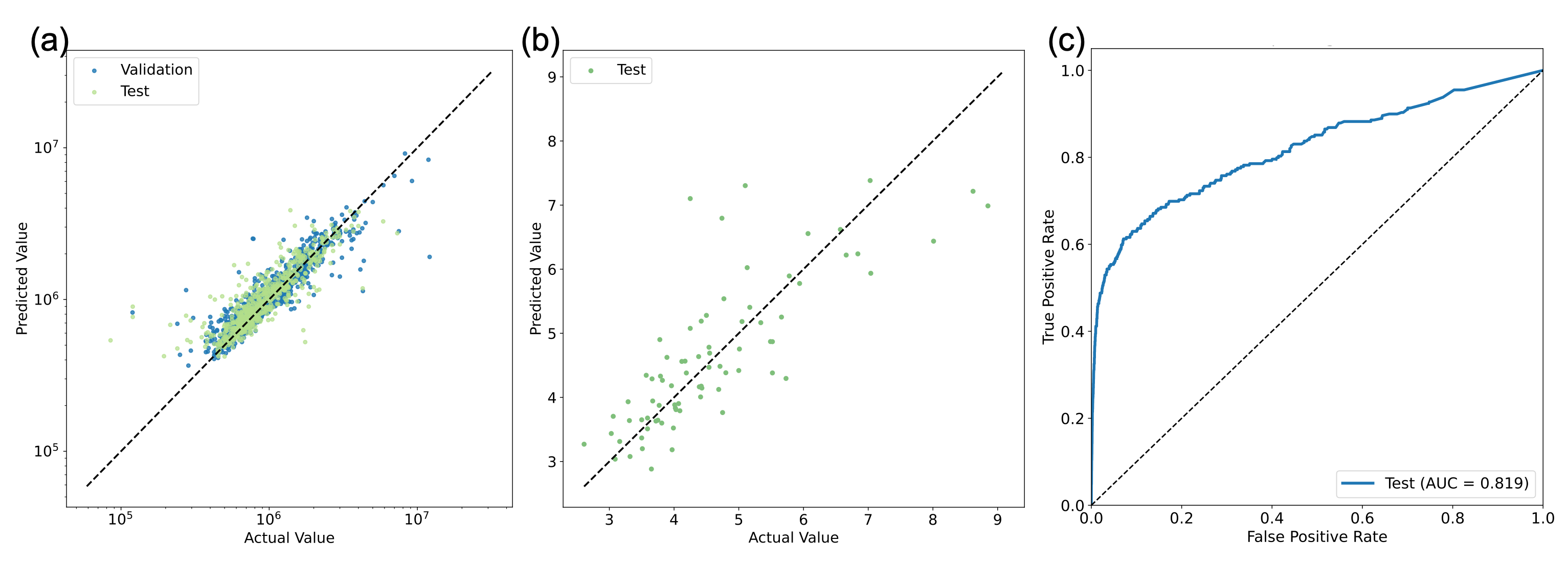}
\caption{\label{Fig4-1}Base ML model performance on three different datasets. (a) King County home price prediction (regression); (b) Polymer dielectric constant prediction (regression); (c) Drug HIV replication inhibition ability prediction (classification).}
\end{figure}


\subsection{Feature Importance}
\label{Appendix-Featureimportance}

\begin{figure}[H]
\centering
\includegraphics[width=0.75\linewidth]{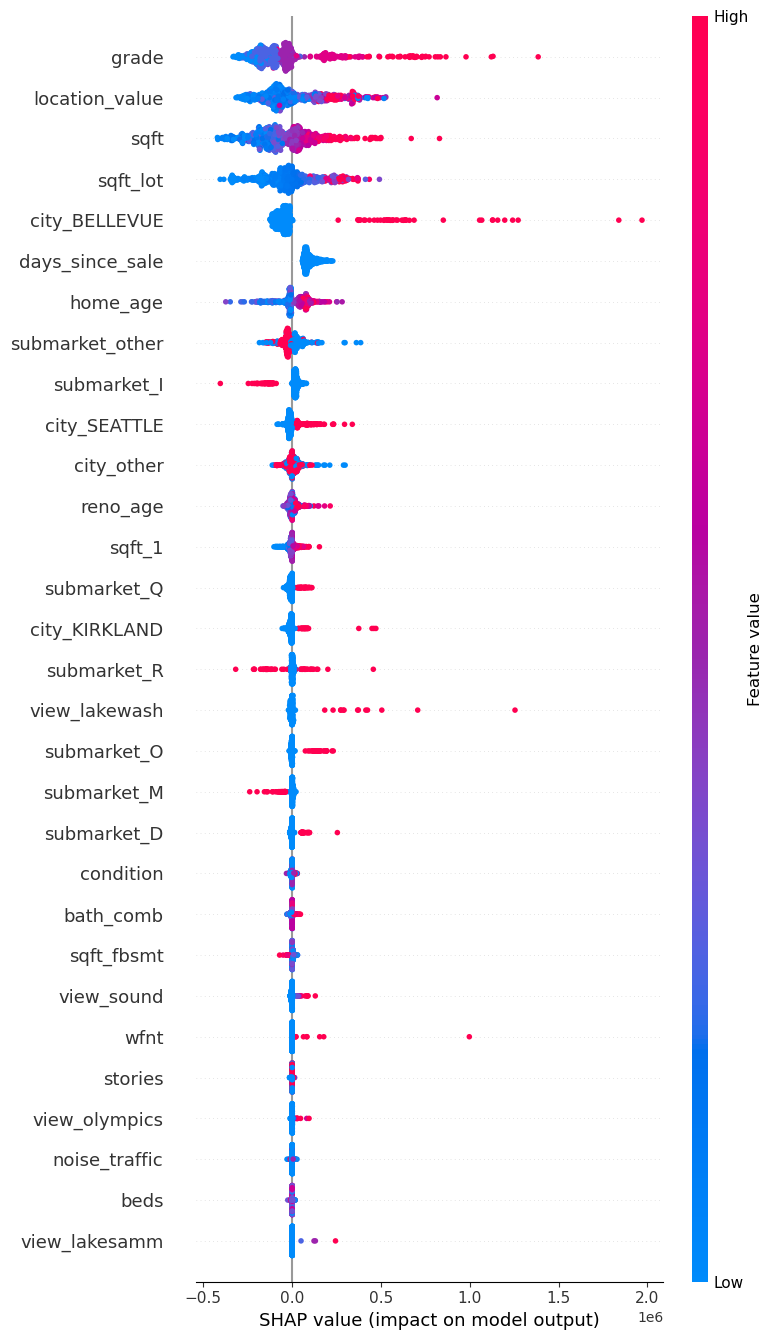}
\caption{\label{Appendix-KingImportance}Beeswarm plot showing the initial SHAP feature importance analysis on the Home dataset. Features are ranked from top to bottom based on their average absolute SHAP values, with higher-ranked features having greater influence on the model's predictions.}
\end{figure}

\begin{figure}[H]
\centering
\includegraphics[width=1\linewidth]{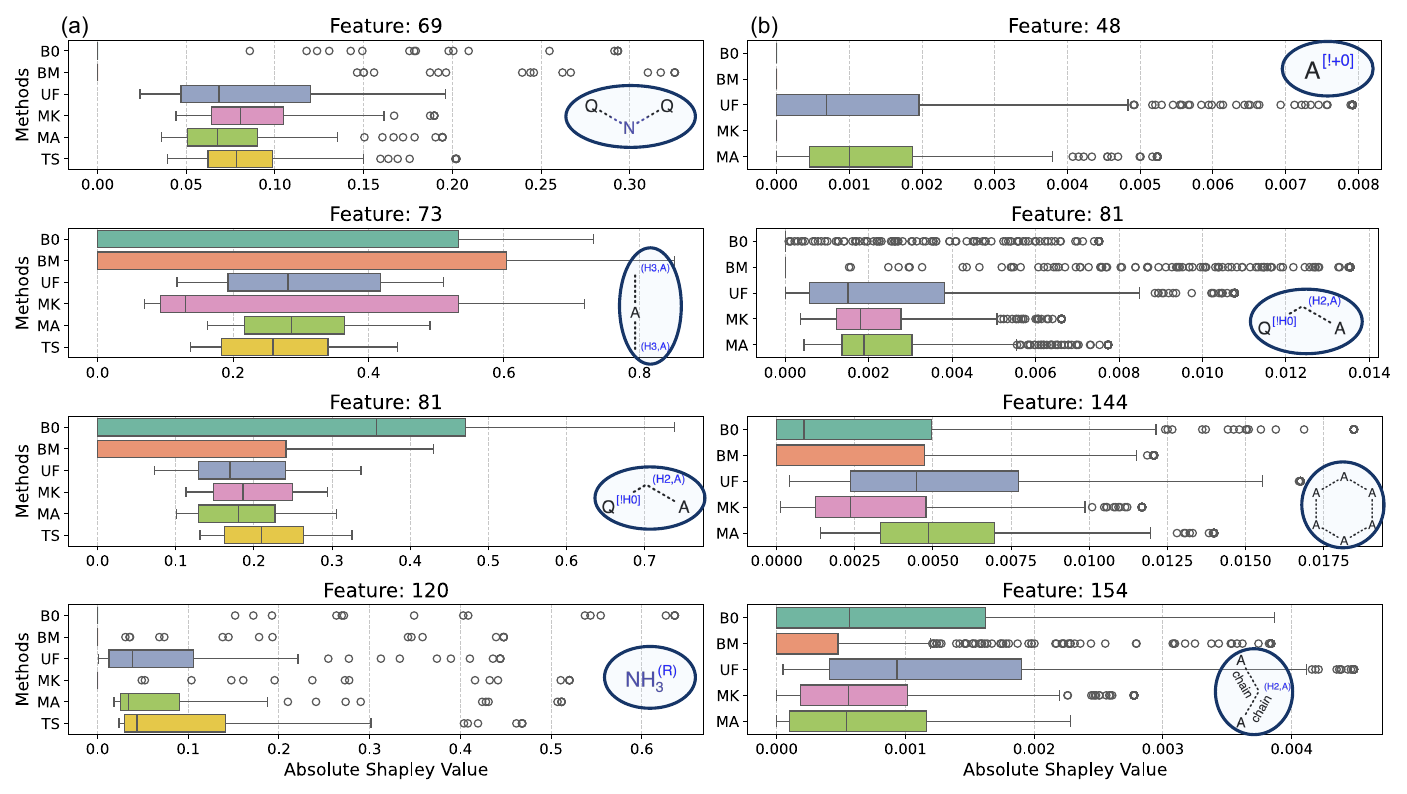}
\caption{\label{Appendix-PolyDrug}Feature importance scores (mean absolute Shapley values) for selected important features of datasets (a) Polymer and (b) Drug. The corresponding molecular structures of each feature are included in each sub-panel.}
\end{figure}

\subsection{Examples of Feature Attribution}
\label{Appendix-FeatureAttributionLocal}
\begin{figure}[H]
\centering
\includegraphics[width=0.8\linewidth]{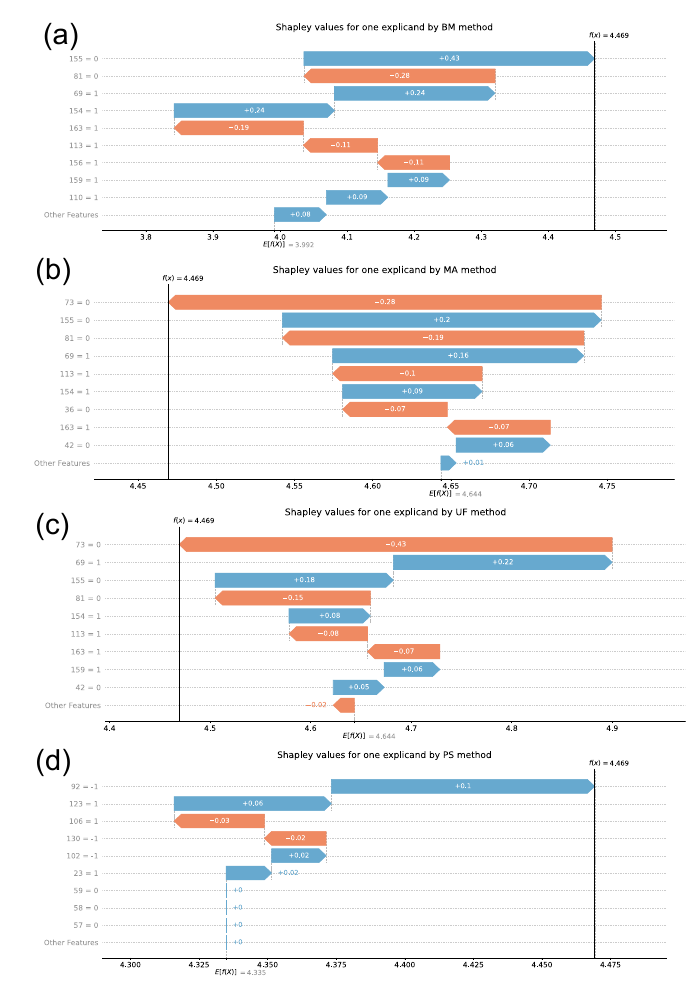}
\caption{\label{append_local_poly}Waterfall plot of feature attribution estimations for an explicand from Polymer dataset using different methods: (a) BM, (b) MA, (c) UF, and (d) PS.}
\end{figure}

\begin{figure}[H]
\centering
\includegraphics[width=0.8\linewidth]{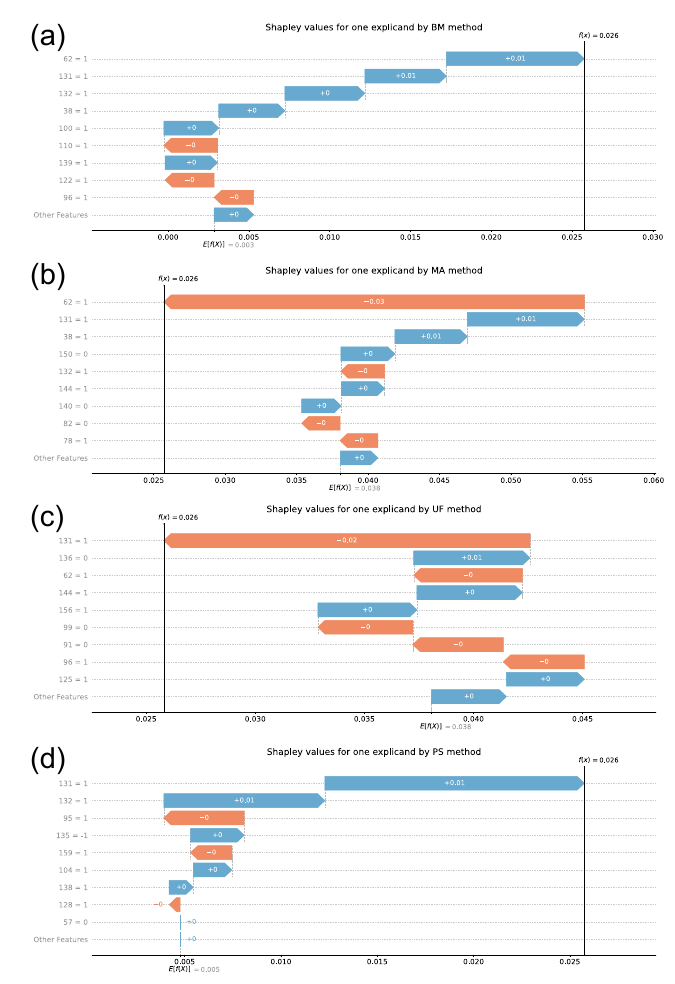}
\caption{\label{append_local_drug}Waterfall plot of feature attribution estimations for an explicand from HIV dataset using different methods: (a) BM, (b) MA, (c) UF, and (d) PS.}
\end{figure}

\section{Guidance for Building an Explainable Model}
\label{Appendix-Protocol}

As ML models become increasingly integral to decision-making in high-stakes domains, the need for model explainability has risen significantly. Ensuring that a model’s predictions are transparent and interpretable is vital for building trust among stakeholders, meeting regulatory requirements, and addressing ethical concerns. This protocol presents a systematic approach to feature auditing for tabular data prior to model construction, emphasizing techniques such as Shapley values to derive feature-level explanations. By following these steps, practitioners can ensure that final models are not only accurate but also interpretable, ethical, and aligned with domain knowledge.

\subsection{Feature Inclusion/Exclusion Criteria}
This protocol defines several criteria to determine whether a feature should be included or excluded in an explainable model. While each project may prioritize these criteria differently, the following list offers a comprehensive framework.

\begin{itemize}
    \item \textbf{Domain Expert Knowledge.} Features with clear domain relevance are more understandable and can make explanations more actionable. For example, a home price prediction model might remove “Improved Tax Value” if “Land Value” already captures the most straightforward assessment of value. Improved Tax Value may be harder to explain due to building-quality variables, whereas Land Value’s drivers (e.g., size, location, zoning) are more transparent.
    \item \textbf{Redundant Features.} If multiple features convey the same information (e.g., \textit{raw} versus \textit{log-transformed}), retaining a single representation simplifies model explanations. Redundant features can inflate model complexity and dilute interpretability. It is often preferable to keep the form that yields the clearest explanations or performs best in preliminary analyses. For example, we might have both “zip\_code” and “GPS\_coordinates” for the same property. If detailed latitude/longitude is already being used, the zip code may become redundant—or vice versa—unless there’s added interpretability or regulatory reasons to keep both.
    \item \textbf{Data Quality.} Exclude or impute features with unreliable, inaccurate, or inconsistent data. Poor-quality data undermines both model performance and the validity of subsequent explanations, rendering Shapley values or other interpretability techniques less reliable.
    \item \textbf{Low Coverage, High Cardinality, and Sparse Features.}. Features with numerous missing values or extremely high cardinality (e.g., more than 10k unique categories) can skew Shapley values unless carefully handled. For example, in our Home dataset, the \textit{subdivision} feature had over 10,000 unique values, making explanations cumbersome. The \textit{present\_use} feature was removed because 92\% of homes fell into a single category (“Single Family Detached”), limiting its contribution to predictive power and interpretability.
    \item \textbf{Ethical and Legal Restrictions.} Exclude or properly handle features tied to sensitive attributes (e.g., race, gender) to comply with regulations and prevent unethical bias. In many jurisdictions, using demographic data for certain decisions is prohibited or heavily regulated, and may also inadvertently introduce discriminatory outcomes.
    \item \textbf{Highly Correlated Features.} Including features that are highly correlated with others can distort the interpretation of Shapley values, as the importance might be unfairly attributed or spread across correlated features. Consider removing or combining these features. Among highly correlated variables, to decide which one to keep, one could use domain knowledge to identify the feature(s) with clearer causal relevance. If domain insight is unavailable, practitioners use preliminary feature importance analyses (e.g., random forest importance, correlation matrix) to retain the most impactful feature; or analyze the cross-validation model performance.
    \item \textbf{No or Low Impact Features.} Features that do not significantly influence predictions can be removed after testing with exploratory data analysis, correlation analysis, or model-based importance metrics. For example, if initial Shapley or permutation importance reveals that a feature (e.g, “view\_otherwater” in our home price prediction use case) contributes negligibly to predictions, it can be excluded, simplifying subsequent explanations.

\end{itemize}

\subsection{Step-by-Step Protocol}
The protocol can be divided into the following steps to ensure a consistent and thorough approach:

\begin{enumerate}
\item \textbf{Initial Data Collection and Preprocessing}
\begin{itemize}
    \item Gather all potential features.
    \item Apply basic cleaning (remove obvious errors, standardize formats, handle missing data).
    \item Document any transformations (e.g., log transformations, scaling) performed.
\end{itemize}

\item \textbf{Collecting Domain Expertise}
\begin{itemize}
    \item Conduct interviews or workshops with domain experts to ascertain which features have clear business or operational relevance.
    \item Flag features that are not easily explainable or contradict known domain relationships.
\end{itemize}

\item \textbf{Exploratory Data Analysis}
\begin{itemize}
    \item Generate descriptive statistics, correlation matrices, and distribution plots.
    \item Identify high-cardinality features, missing-value patterns, or candidate transformations (e.g., binning continuous variables).
\end{itemize}

\item \textbf{Feature Reduction and Preliminary Importance Scoring}
\begin{itemize}
    \item Use methods such as correlation thresholds, variance-based feature selection, or tree-based feature importance to shortlist candidate features.
    \item Document any decisions to exclude features at this stage.
\end{itemize}

\item \textbf{Ethical and Legal Review}
\begin{itemize}
    \item Ensure compliance with relevant laws and ethical standards by removing or anonymizing sensitive attributes.
    \item If neccessary, consult legal experts or ethics committees for final approval.
\end{itemize}

\item \textbf{Iterative Explainable Feature Refinement}
\begin{itemize}
    \item Train a preliminary model (e.g., gradient boosting, random forest).
    \item Apply explainability methods (e.g., SHAP) to assess each feature’s marginal contribution.
    \item Remove or modify features exhibiting minimal importance or interpretable value.
    \item Repeat iteratively until a stable set of informative, ethically sound, and interpretable features emerges (see Fig. \ref{iterative_feature_refinement}).
\end{itemize}

\begin{figure}[H]
\centering
\includegraphics[width=1\linewidth]{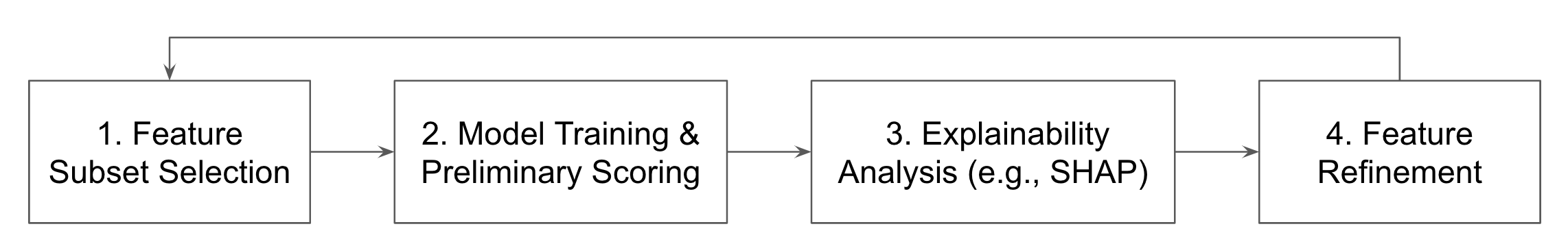}
\caption{\label{iterative_feature_refinement}An iterative workflow for feature selection and refinement in building explainable machine learning models. The process involves (1) selecting a subset of features from the previous steps in the protocol, (2) training and scoring a preliminary model, (3) conducting explainability analysis using methods like SHAP, and (4) refining features based on insights from the analysis.}
\end{figure}

\item \textbf{Final Feature Set and Documentation}
\begin{itemize}
    \item Produce a comprehensive record of features retained, rationale for exclusion, and data preprocessing steps.
    \item Ensure reproducibility by version-controlling code and storing raw data.
    \item This final set forms the basis for the deployment model used in decision-making.
\end{itemize}

\end{enumerate}

By following this protocol, practitioners can systematically evaluate and refine a robust feature set tailored for explainable machine learning models. Incorporating domain expertise, data quality checks, redundancy screening, ethical considerations, and iterative refinement processes ensure that final models maintain both high predictive performance and meaningful interpretability. This structured approach not only enhances transparency and trustworthiness of ML solutions in high-stakes domains but also supports efficient business operations and adherence to regulatory frameworks.

\end{document}